\documentclass[runningheads]{llncs}
\usepackage[T1]{fontenc}
\usepackage{graphicx}
\usepackage{hyperref}

\usepackage{color}

\newcommand*{\titletext}{Using Abstraction for Interpretable Robot Programs in Stochastic Domains}
\hypersetup{
  pdftitle = {\titletext},
  pdfauthor = {Till Hofmann, Vaishak Belle}
}

\usepackage{acronym}
\acrodef{BAT}{basic action theory}
\newacroplural{BAT}{basic action theories}
\acrodef{HTN}{hierarchical task network}
\acrodef{oi}{observational indistinguishability}

\usepackage{amsmath}
\usepackage{amssymb}
\usepackage{mleftright}
\usepackage{xspace}
\usepackage{siunitx}

\newcommand*{\golog}{\textsc{Golog}\xspace}

\newcommand*{\allegro}{\textsc{Allegro}\xspace}

\newcommand*{\es}{\ensuremath{\mathcal{E \negthinspace S}}\xspace}

\newcommand*{\ds}{\texorpdfstring{\ensuremath{\mathcal{D \negthinspace S}}}{DS}\xspace}

\newcommand*{\poss}{\ensuremath\mathit{Poss}}

\newcommand*{\eqdef}{\ensuremath{:=}}

\newcommand*{\equivspace}{\ensuremath\,\equiv\;}

\newcommand*{\belconnector}{\ensuremath{\mathbf{B}}\xspace}

\newcommand*{\bel}[2]{\ensuremath{\belconnector\mleft(#1\,\mathbf{:}\,#2\mright)}}

\newcommand*{\know}[1]{\ensuremath{\mathbf{Know}(#1)}}

\newcommand*{\tnext}[2][]{\ensuremath{\mathbf{X}^{#1}_{#2}}\if#1{\,}\fi}
\newcommand*{\tprev}[1]{\ensuremath{\mathbf{V}_{#1}}\if#1{\,}\fi}
\newcommand*{\fut}[1]{\ensuremath{\mathbf{F}_{#1}}\if#1{\,}\fi}
\newcommand*{\past}[1]{\ensuremath{\mathbf{P}_{#1}}\if#1{\,}\fi}
\newcommand*{\glob}[1]{\ensuremath{\mathbf{G}_{#1}}\if#1{\,}\fi}
\newcommand*{\hist}[1]{\ensuremath{\mathbf{H}_{#1}}\if#1{\,}\fi}
\newcommand*{\mi}[1]{\ensuremath{\mathit{#1}}}
\newcommand*{\la}{\left\langle}
\newcommand*{\ra}{\right\rangle}
\mathchardef\mhyphen="2D % Define a "math hyphen"

\newcommand*{\near}{\ensuremath{\mathit{near}}\xspace}
\newcommand*{\far}{\ensuremath{\mathit{far}}\xspace}
\newcommand*{\goto}{\ensuremath{\mathit{goto}}\xspace}
\newcommand*{\sigmah}{\ensuremath{\Sigma_{\mi{goto}}}\xspace}
\newcommand*{\sigmal}{\ensuremath{\Sigma_{\mi{move}}}\xspace}
\newcommand*{\loc}{\ensuremath{\mathit{Loc}}\xspace}
\newcommand*{\at}{\ensuremath{\mathit{At}}\xspace}
\newcommand*{\move}{\ensuremath{\mathit{move}}\xspace}

\newcommand*{\sonar}{\ensuremath{\mathit{sonar}}\xspace}

\newcommand*{\gwhile}{\ensuremath{\;\mathbf{while}\;}}
\newcommand*{\gdo}{\ensuremath{\;\mathbf{do}\;}}
\newcommand*{\gdone}{\ensuremath{\;\mathbf{done}\;}}
\newcommand*{\gif}{\ensuremath{\;\mathbf{if}\;}}
\newcommand*{\gthen}{\ensuremath{\;\mathbf{then}\;}}
\newcommand*{\gfi}{\ensuremath{\;\mathbf{fi}\;}}
\newcommand*{\gelif}{\ensuremath{\;\mathbf{elif}\;}}

\newcommand*{\stateset}{\ensuremath{S}\xspace}

\newcommand*{\oi}{\ensuremath{\mathit{oi}}}

\newcommand*{\wlfull}[2]{\ensuremath{\stateset^{#1}_{\ifthenelse{\equal{#2}{\true}}{#2}{m\mleft(#2\mright)}}}}

\newcommand*{\lb}{\mleft \lbrack}
\newcommand*{\rb}{\mright \rbrack}

\newcommand*{\true}{\textsc{True}\xspace}

\begin{document}
\title{\titletext}
%
%\titlerunning{Abbreviated paper title}
% If the paper title is too long for the running head, you can set
% an abbreviated paper title here
%
\author{Till Hofmann\inst{1} \and
Vaishak Belle\inst{2}
}
\authorrunning{T.~Hofmann, V.~Belle}
% First names are abbreviated in the running head.
% If there are more than two authors, 'et al.' is used.
%
\institute{
  Knowledge-Based Systems Group, RWTH Aachen University, Aachen, Germany\\
  \email{hofmann@kbsg.rwth-aachen.de} \and
  University of Edinburgh, Edinburgh, UK\\
  \email{vbelle@ed.ac.uk}
}
\maketitle              % typeset the header of the contribution
\begin{abstract}
  A robot's actions are inherently stochastic, as its sensors are noisy and its actions do not always have the intended effects.
  For this reason, the agent language \golog has been extended to models with degrees of belief and stochastic actions.
  %allow for stochastic actions and sensors that cannot capture the full world state.
  While this allows more precise robot models, the resulting programs are much harder to comprehend, because they need to deal with the noise, e.g., by looping until some desired state has been reached with certainty,
  and because the resulting action traces consist of a large number of actions cluttered with sensor noise.
  %Additionally, they produce execution traces that are cluttered with sensor noise and stochastic action results, and which consist of a large number of actions, even for simple programs.
  To alleviate these issues, we propose to use \emph{abstraction}.
  We define a high-level and nonstochastic model of the robot and then map the high-level model into the lower-level stochastic model.
  The resulting programs are much easier to understand, often do not require belief operators or loops, and produce much shorter action traces.
  %At the same time, by translating the program to the lower-level model, we may still deal with the robot's stochastic actions.
%\keywords{Agent Programs  \and Stochastic Actions \and Abstraction.}
\end{abstract}

\section{Introduction}

Classical approaches to model robot behavior such as
%task planning~\cite{ghallabAutomatedPlanningTheory2004} or
\golog~\cite{levesqueGOLOGLogicProgramming1997} assume complete knowledge of the word state as well as deterministic actions.
%that the world is completely known to the agent and that the robot's actions are deterministic.
However, both assumptions are often violated on real robots:
the robot's sensor cannot completely capture the world, thus requiring some way to represent incomplete knowledge,
and a robot's sensors and effectors are inherently noisy, necessitating a model of stochastic actions.
%, a robot sensor is never exact and an action may fail or have different results.
Consider a simple robot equipped with a sonar sensor that is driving towards a wall (inspired from \cite{bacchusReasoningNoisySensors1999}).
The sensor reading is imprecise and may produce incorrect sensor reading.
%read a distance of $\SI{2}{\metre}$ while it is actually $\SI{3}{\metre}$ away.
Additionally, when doing a \move action, the robot may get stuck with some probability.
%when doing a \move action, the robot may get stuck with some probability and therefore not move at all.
%To model such systems, the situation calculus~\cite{reiterKnowledgeActionLogical2001} has been extended with stochastic actions and degrees of belief~\cite{bacchusReasoningNoisySensors1999,belleReasoningProbabilitiesUnbounded2017}.
One approach to model such systems is an extension of the situation calculus~\cite{reiterKnowledgeActionLogical2001} with stochastic actions and degrees of belief~\cite{bacchusReasoningNoisySensors1999,belleReasoningProbabilitiesUnbounded2017}.
The core idea is to model the agent's belief with possible worlds, where each world has a certain weight, defining the probability that this world is the actual world.
This allows to model degrees of belief, e.g., saying ``the robot beliefs to be $\SI{2}{\metre}$ away with certainty $0.5$''.
An action may then possibly have several outcomes, each specified with some likelihood.
%As an example, if the robot tries to execute $\move(1)$, it may actually move by $\SI{1}{\metre}$ with a probability of $.8$, or it may move with an error of $\pm \SI{1}{\metre}$ with a probability of $.1$ each.

While there has been progress on reasoning about belief-based programs~\cite{liuReasoningBeliefsMetabeliefs2021} and programming languages such as \allegro~\cite{belleALLEGROBeliefbasedProgramming2015} allow to write belief-based programs, including stochastic actions in the model has the disadvantage that the resulting programs become significantly harder to understand.
Dealing with noisy actions often requires many actions and loops to reach a desired state with certainty.
Additionally, such stochastic programs often have many possible traces, making it more difficult to interpret one particular run of a program.

In this paper, we illustrate how \emph{abstraction} can be used to deal with these issues.
%To deal with these issues, we propose to use \emph{abstraction}.
Extending on abstraction of \acp{BAT} in the classical situation calculus~\cite{banihashemiAbstractionSituationCalculus2017}, we propose to use abstractions of stochastic domains, where the resulting abstract \ac{BAT} does not contain any noisy sensors or effectors and is therefore nonstochastic.
The resulting programs are easier to write and understand and the resulting traces are free of stochastic effects, thus making them much easier to comprehend.

The remainder of the paper is structured as follows~\footnote{In this paper, we focus on motivating the use of abstraction for interpretable programs. We refer to \cite{hofmannAbstractingNoisyRobot2022} for a full technical discussion.}:
In \autoref{sec:low-level-program}, we introduce belief-based programs based on the logic \ds~\cite{belleReasoningProbabilitiesUnbounded2017}.
%Rather than providing the full semantics,
We present an example for the robot described above and we show how even simple programs induce traces that are non-trivial to understand.
In \autoref{sec:abstraction}, we define a more abstract and nonstochastic model that can be mapped to the lower-level model, resulting in more comprehensible programs.
%In contrast to the low-level program, the abstract program clearly expresses its intent and induces trivial traces.
We conclude in \autoref{sec:conclusion}.

%\begin{itemize}
%  \item robots typically have noisy sensors and actuators
%  \item recent interest in representing noisy sensors and actuators by extending knowledge-based programs to belief-based programs~\cite{bacchusReasoningNoisySensors1999,belleReasoningProbabilitiesUnbounded2017,belleALLEGROBeliefbasedProgramming2015,liuProgressionBelief2021,liuReasoningBeliefsMetabeliefs2021}
%  \item here: degrees of belief, e.g., the robot knows with certainty $0.5$ that it is near the wall
%  \item while there has been some progress on reasoning about belief-based programs~\cite{liuProgressionBelief2021,liuReasoningBeliefsMetabeliefs2021}, they are notoriously hard to understand
%  \item solution: abstract the belief-based program to a deterministic knowledge-based program
%  \item sub-programs that deal with noise are abstracted away by a single action
%  \item resulting traces are easier to understand and therefore provide a better explanation for why the agent decided to do certain actions
%\end{itemize}

\section{Belief-Based Programs with Stochastic Actions}\label{sec:low-level-program}

\ds~\cite{belleReasoningProbabilitiesUnbounded2017} is a modal variant of the situation calculus with degrees of belief and stochastic actions.
%Put differently, it extends the logic \es~\cite{lakemeyerSemanticCharacterizationUseful2011}, which is a modal variant of the situation calculus, with degrees of belief.
Similar to \es~\cite{lakemeyerSemanticCharacterizationUseful2011}, it is based on a possible worlds semantics, where worlds are part of the semantics and do not occur as terms in the language.
In \ds, all worlds have a weight, which defines the probability of each world being the true world.
%More precisely, an epistemic state in \ds consists of a set of distributions, where each distribution assigns a weight to every possibe world.
The modal operator $\bel{\alpha}{r}$ expresses that $\alpha$ is believed with degree $r$, e.g., $\bel{\loc(2)}{0.5}$ expresses that the agent believes ``the distance to the wall is \SI{2}{\metre}'' with degree $0.5$.
We also write $\know{\alpha}$ for $\bel{\alpha}{1}$.
%to express that the agent knows $\alpha$ with certainty.

\ds has two action modalities: $[a]$ and $\square$, where $[a]\alpha$ is to be read as ``$\alpha$ holds after doing action $a$'' and $\square \alpha$ is to be read as ``$\alpha$ holds after any sequence of actions''.
As in the situation calculus, a \ac{BAT} defines a domain by axiomatizing the initial situation, action preconditions, and effects.
Additionally, noisy actions are modeled with the \emph{action likelihood} $l$, where $l(a, u)$ expresses that the likelihood of action $a$ is $u$.
After doing an action, the agent may not always distinguish which instance of the action has actually been executed.
This is expressed with \acfi{oi} axioms, where $\oi(a, a')$ expresses that the agent cannot distinguish the actions $a$ and $a'$.

We do not present a full account of \ds, but rather present an example.
We model the robot's movement with the action $\move(x, y)$, where $x$ is the intended distance and $y$ is the distance that the robot actually moved, and
with a single fluent predicate $\loc(x)$ that describes the position of the robot.
A \ac{BAT} \sigmal defining the scenario from above may look as follows:
\begin{itemize}
  \item After doing action $a$, the robot is at position $x$ if $a$ is a \move action that moves the robot to location $x$, if $a$ is a \sonar action that measures distance $x$, or if $a$ is neither of the two actions and the robot was at location $x$ before%
    \footnote{We assume that free variables are universally quantified from the outside and that $\square$ has lower syntactic precedence than the logical connectives, so that $\square[a]\loc(x) \equiv \gamma$ stands for $\forall a.\, \square \left([a] \loc(x) \equiv \gamma\right)$.}:
    \begin{align*}
      \square \lb a \rb \loc(x) \equivspace
      &\exists y,z, \left(a = \move(y,z) \wedge \loc(l) \wedge x = l + z \right)
      \vee a = \sonar(x)
      \\
      & \vee \neg \exists y, z \left(a = \move(y,z) \vee a = \sonar(y)\right) \wedge \loc(x)
    \end{align*}
  \item
    A \move action is possible if the robot moves either one step to the back or to the front.
    A \sonar action is always possible:
    \begin{align*}
      \square \poss(a) \equivspace
      & \exists x,y \left(a = \move(x,y) \wedge \left(x = 1 \vee x = -1\right)\right)
      \vee \exists z \left(a = \sonar(z)\right)
    \end{align*}
  \item Action likelihood axioms: For the \sonar action, the likelihood that the robot measures the correct distance is $0.8$, the likelihood that it measures a distance with an error of $\pm 1$ is $0.1$.
    Furthermore, for the \move action, the likelihood that the robot moves the intended distance $x$ is $0.6$, the likelihood that the actual movement $y$ is off by $\pm 1$ is $0.2$:
    \begin{align*}
      \square l(a, u) \equivspace
      & \exists z \left( a = \sonar(z) \wedge \loc(x) \wedge u = \Theta(x, z, .8, .1)\right)
      \\
      &  \vee \exists x,y \left(a = \move(x, y) \wedge u = \Theta(x, y, .6, .2)\right)
      \\
      &  \vee \neg \exists x,y,z \left(a = \move(x,y) \vee a = \sonar(z)\right) \wedge u = .0
    \end{align*}
    where $\Theta(u,v,c,d) =
      \begin{cases} c & \text{ if } u = v \\ d & \text{ if } |u-v| = 1 \\ 0 & \text{ otherwise} \end{cases}$.
  \item  The robot cannot detect the distance that it has actually moved, i.e., any two actions $\move(x, y)$ and $\move(x, z)$ are o.i.:
    %For the \sonar action, the robot can always distinguish the action from other actions:
    \begin{align*}
      \square \oi(a, a') \equivspace
            & \exists x, y, z \left(a = \move(x, y) \wedge a' = \move(x, z)\right)
      \vee a = a'
    \end{align*}
  \item Initially, the robot is $\SI{3}{\metre}$ away from the wall: $\forall x (\loc(x) \equivspace x = 3)$
\end{itemize}

%\subsection{A \golog Program to Move the Robot}
Based on this \ac{BAT}, we define a program that first moves the robot close to the wall and then back\footnote{The unary $\move(x)$ can be understood as abbreviation $\move(x) \eqdef \pi y\, \move(x, y)$, where nature nondeterministically picks the distance $y$ that the robot really moved (similarly for $\sonar()$).}:
%such that it ends far away from the wall:
\begin{align*}
   \sonar();
  & \gwhile \neg \know{\exists x \left(\loc(x) \wedge x \leq 2\right)}
  \gdo \move(-1); \sonar() \gdone;
  \\ &\gwhile \neg \know{\exists x \left(\loc(x) \wedge x > 5\right)}
  \gdo \move(1); \sonar() \gdone
\end{align*}

The robot first measures its distance to the wall and then moves closer until it knows that its distance to the wall is less than $\SI{2}{\metre}$.
Afterwards, it moves away until it knows that is more than $\SI{5}{\metre}$ away from the wall.
%It then starts moving away and continues doing so until it knows that is more than $\SI{5}{\metre}$ away from the wall.
As the robot's \move action is noisy, each \move is followed by \sonar to measure how far it is away from the wall.
One possible execution trace of this program may look as follows:
\begin{align}
  z_l = \langle &\sonar(3), \move(-1, 0), \sonar(3), \move(-1, -1),
   \sonar(2), \move(-1, -1), \nonumber
  \\
  & \sonar(1), \move(1, 1), \sonar(3), \move(1, 1),
  \sonar(2), \move(1, 1), \nonumber
  \\ &
  \sonar(4), \move(1, 0),
  \sonar(4), \move(1, 1), \sonar(6) \label{eqn:low-level-trace}
  \rangle
\end{align}
First, the robot (correctly) senses that it is $\SI{3}{\metre}$ away from the wall and starts moving.
%towards the wall.
%Next, the robot moves towards the wall.
However, the first \move does not have the desired effect: the robot intended to move by $\SI{1}{\metre}$ but actually did not move (indicated by the second argument being $0$).
After the second \move, the robot is  at $\loc(2)$, as it started at $\loc(3)$ and moved successfully once.
However, as its sensor is noisy and it measured $\sonar(2)$, it believes that it could also be at $\loc(3)$.
For safe measure, it executes another \move and then senses $\sonar(1)$, after which it knows for sure that it is at a distance $\leq \SI{2}{\metre}$.
%The robot continues to move and eventually senses $\sonar(1)$, after which it knows for sure that it is at a distance $\leq \SI{2}{\metre}$.
In the second part, the robot moves back until it knows that it has reached a distance $> \SI{5}{\metre}$.
%For the second part, we can observe that even though it moved successfully away from the wall ($\move(1, 1)$), its sensor actually says that it moved towards the wall (first $\sonar(3)$ and then $\sonar(2)$).
%This is possible because the first measurement was off by $+1$, while the second measurement was off by $-1$, thus being consistent with \sigmal.
%Similar to before, we can see that the robot continues moving away until it knows for sure that it has reached a distance $x > 5$.
As this simple example shows, the trace $z_l$ is already quite hard to understand.
While it is clear from the \ac{BAT} what each action does, the robot's intent is not immediately obvious and the trace is cluttered with noise and low-level details.

%\begin{itemize}
%  \item in \ds, it is possible to express degrees of belief, e.g., we can say that the robot believes ``the distance to the wall is 2 units'' with degree $0.5$
%  \item modal variant of the situation calculus, based on \es
%  \item more precisely, a modal variant of an extension of the situation calculus with degrees of belief~\cite{bacchusReasoningNoisySensors1999}
%  \item as in \es, standard names $\rightarrow$ quantification is understood substitutionally
%  \item modal operator $\bel{\alpha}{r}$ expresses that $\alpha$ is believed with degree $r$
%  \item possible world semantics with weighted worlds
%  \item an epistemic state consists of a set of distributions, where each distribution assigns a weight to a world
%  \item a \ac{BAT} defines a domain and consists of:
%\end{itemize}

\section{Using Abstraction to Obtain Interpretable Programs}\label{sec:abstraction}
To hide away the low-level details such as noisy \move actions, we propose to use \emph{abstraction}.
Similar to \cite{banihashemiAbstractionSituationCalculus2017}, abstraction in \ds is a mapping of a high-level \ac{BAT} to a low-level \ac{BAT}.
%To obtain such a mapping,
We first define the high-level \ac{BAT} and then map each fluent and action to the lower level.
Continuing our example, we can define a second, high-level \ac{BAT} that consists of the locations \near and \far, the fluent \at that specifies the current location of the robot, and the single action \goto, which is an idealized \move action without noise.
The high-level \ac{BAT} \sigmah looks as follows:
\begin{itemize}
  \item After doing action $a$, the robot is at location $l$ if $a$ is the action $\goto(l)$ or if $a$ is no \goto action and the robot has been at $l$ before:
    \[
      \square \lb a \rb \at(l) \equivspace
      a = \goto(l) \vee \neg \exists x \left(a = \goto(x)\right) \wedge \at(l)
    \]
  \item The robot may do action $a$ if $a$ is a \goto action to a valid location:
    \[
      \square \poss(a) \equivspace
      a = \goto(\near) \vee a = \goto(\far)
    \]
  \item The \goto action is not noisy: %, thus the action likelihood axiom says that there is exactly one outcome with likelihood $1.0$:
    \begin{align*}
      \square &l(a, u) \equivspace \exists x \left(a = \goto(x)\right) \wedge u = 1.0
      \vee \neg \exists x \left(a = \goto(x)\right) \wedge u = 0.0
    \end{align*}
  \item The agent can distinguish all actions: $\square \oi(a, a') \equivspace a = a'$
  \item Initially, the robot is at the location \near: $\forall l ( \at(l) \equivspace l = \near )$
\end{itemize}

Next, we define a \emph{refinement mapping} that maps each high-level fluent to a low-level formula and each high-level action to a low-level program:
%In our case, the mapping may look as follows:
\begin{itemize}
  \item The high-level fluent $\at(l)$ is mapped to a low-level formula by translating the distance to the two locations $\near$ and $\far$:
    \begin{align*}
      \at(l) \mapsto &\big(
      %  l = \infront \wedge \exists x \left(\loc(x) \wedge x \leq 2\right)
      %\\
      %& \vee
      l = \near \wedge \exists x \left(\loc(x) \wedge x \leq 2\right)
       \vee l = \far \wedge \exists x \left(\loc(x) \wedge x > 5\right)\big)
    \end{align*}
  \item The action $\goto$ is mapped to a program that guarantees that the robot reaches the right position:
    \begin{align*}
       \goto(&x) \mapsto
                \sonar();
                \\
                & \gif x = \near \gthen \\
                & \quad \gwhile \neg \know{\exists x \left(\loc(x) \wedge x \leq 2\right)}
                \gdo \move(-1); \sonar() \gdone
                \\
                & \gelif x = \far \gthen \\
                & \quad \gwhile \neg \know{\exists x \left(\loc(x) \wedge x > 5\right)}
                \gdo \move(1); \sonar() \gdone; \gfi
    \end{align*}
%As we can see, when doing $\goto(x)$, the robot continues moving forward or backward until it knows that it has reached a position in the desired region.
%To do so, it interleaves single-step move actions ($\move(1)$ or $\move(-1)$) with \sonar actions.
%As the sonar is off by a distance of at most $\pm 1$, it is always able to reach the goal region with certainty.
\end{itemize}

Using the refinement mapping, we can translate a high-level program based on the \ac{BAT} \sigmah to a low-level program based on the \ac{BAT} \sigmal.
As \sigmah does not contain any sensing or noisy actions, we obtain a nonstochastic program that is much simpler to understand than the low-level program shown above.
The program that first moves close to the wall and then moves back only needs two actions and does not require any belief operators:
\[
  \goto(\near); \goto(\far)
\]
Note that when the program is executed, each $\goto$ action is translated into a corresponding low-level program, as defined by the mapping above.
However, from the high-level perspective, the program only allows a single trace
$ z_h = \la \goto(\near), \goto(\far) \ra$.
Compared to the low-level trace $z_l$ from \autoref{eqn:low-level-trace}, $z_h$ is much easier to understand.
It only consists of two actions, which are exactly the two actions from the program.
It does not contain any noise, which is the reason why the trace is also unique.
Furthermore, we were able to abstract away all sensing actions, further simplifying the resulting traces.

\section{Conclusion}\label{sec:conclusion}

In this paper, we demonstrated how \emph{abstraction} can be used to map a low-level \golog program with stochastic actions to a high-level program that is nonstochastic, does not require any sensing or belief operators, and thus is much easier to understand.
While this requires some additional work to define the mapping between the two models, the mapping is not specific for a given program and thus can be re-used for other programs in the same domain.

% Fakesection Appendix
\subsubsection{Acknowledgements}

T.~Hofmann was partly supported by the Deutsche Forschungsgemeinschaft (DFG, German Research Foundation) -- 2236/1 and the EU ICT-48 2020 project TAILOR (No.~952215).
Part of this work was created during a research visit of T.~Hofmann at the University of Edinburgh, which was funded by the German Academic Exchange Service (DAAD).
V.~Belle was partly supported by a grant from the UKRI Strategic Priorities Fund to the UKRI Research Node on Trustworthy Autonomous Systems Governance and Regulation (EP/V026607/1, 2020-2024).
He was also supported by a Royal Society University Research Fellowship.

%
% ---- Bibliography ----
%
% BibTeX users should specify bibliography style 'splncs04'.
% References will then be sorted and formatted in the correct style.
%
\bibliographystyle{splncs04}
\bibliography{abstraction}
\end{document}